\documentclass[journal]{IEEEtran}

\usepackage{cite}
\usepackage{booktabs}
\usepackage{amsmath,amssymb,amsfonts}
\usepackage{algorithmic}
\usepackage{graphicx}
\usepackage{subcaption}   
\usepackage{textcomp}
\usepackage{comment}
\usepackage{makecell}
\usepackage{multirow}
\usepackage{cite}
\usepackage{setspace}
\usepackage{amsmath}
\usepackage[table,xcdraw]{xcolor}
\usepackage[colorlinks=true,allcolors=blue]{hyperref}
\def\BibTeX{{\rm B\kern-.05em{\sc i\kern-.025em b}\kern-.08em
    T\kern-.1667em\lower.7ex\hbox{E}\kern-.125emX}}
\ifCLASSINFOpdf
\else
\fi

\begin{document}
\title{Reinforcement Learning for Follow-the-Leader Robotic Endoscopic Navigation via Synthetic Data}

\author{Sicong Gao, Chen Qian, Laurence Xian,  Liao Wu,  Maurice Pagnucco, and~Yang~Song$^*$
\and
\thanks{{$^*$}Corresponding author: Yang Song, {\tt\small yang.song1@unsw.edu.au}.}
\thanks{Sicong Gao, Maurice Pagnucco,and Yang Song are with the School
of Computer Science and Engineering, The University of New South Wales,
Sydney 2052, Australia (e-mail: sicong.gao@student.unsw.edu.au; morri@cse.unsw.edu.au; yang.song1@unsw.edu.au)}
\thanks{Chen Qian, Laurence Xian and Liao Wu are with the School of Mechanical and Manufacturing Engineering, The University of New South Wales, Sydney, NSW 2052, Australia. (e-mail: chen.qian1@unsw.edu.au; l.xian@student.unsw.edu.au; liao.wu@unsw.edu.au)}
}

\maketitle


\begin{abstract}
Autonomous navigation is crucial for both medical and industrial endoscopic robots, enabling safe and efficient exploration of narrow tubular environments without continuous human intervention, where avoiding contact with the inner walls has been a longstanding challenge for prior approaches.
We present a follow-the-leader endoscopic robot based on a flexible continuum structure designed to minimize contact between the endoscope body and intestinal walls, thereby reducing patient discomfort. 
To achieve this objective, we propose a vision-based deep reinforcement learning framework guided by monocular depth estimation. 
A realistic intestinal simulation environment was constructed in \textit{NVIDIA Omniverse} to train and evaluate autonomous navigation strategies. Furthermore, thousands of synthetic intraluminal images were generated using \textit{NVIDIA Replicator} to fine-tune the \textit{Depth Anything} model, enabling dense three-dimensional perception of the intestinal environment with a single monocular camera. Subsequently, we introduce a geometry-aware reward and penalty mechanism to enable accurate lumen tracking. Compared with the original Depth Anything model, our method improves $\delta_{1}$ depth accuracy by 39.2\% and reduces the navigation J-index by 0.67 relative to the second-best method, demonstrating the robustness and effectiveness of the proposed approach.

\end{abstract}



\begin{IEEEkeywords}
Reinforcement Learning, Endoscope, Depth Estimation, Autonomous Navigation.
\end{IEEEkeywords}

\IEEEpeerreviewmaketitle

\section{Introduction}

Colonoscopy plays a vital role in the early detection and treatment of colorectal diseases and remains the gold standard for diagnosing colorectal cancer, polyps, and inflammatory bowel conditions. Early screening reduces colorectal cancer mortality by 22\%–33\% \cite{segnan2011once}. Despite its clinical efficacy, traditional colonoscopy relies heavily on manual manipulation, requiring skilled endoscopists to guide a long, flexible endoscope through a narrow and tortuous lumen. In regions with sharp bends, greater insertion force is often required, generating both lateral and longitudinal tension on the colonic wall, which limits its elasticity and increases the risk of mechanical stress \cite{hong2023review}. Such manual operation is labor-intensive and time-consuming, often causing patient discomfort, mucosal injury, or incomplete examination, particularly in anatomically complex segments. Studies have reported that 79.2\% of sedated and 88.7\% of non-sedated patients experience pain or discomfort during the procedure \cite{aljebreen2014sedated}. Moreover, navigation performance strongly depends on the operator’s skill, leading to inconsistent diagnostic quality across examinations.

\begin{figure}[t]
  \centering
  \begin{subfigure}[t]{0.32\columnwidth}
    \centering
    \includegraphics[width=\linewidth]{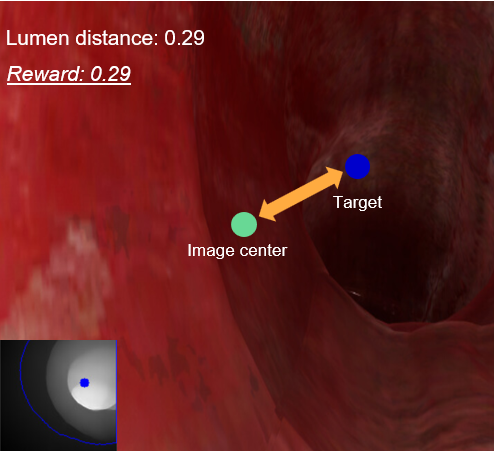}
    \caption{}
    \label{fig:1a}
  \end{subfigure}\hfill
  \begin{subfigure}[t]{0.32\columnwidth}
    \centering
    \includegraphics[width=\linewidth]{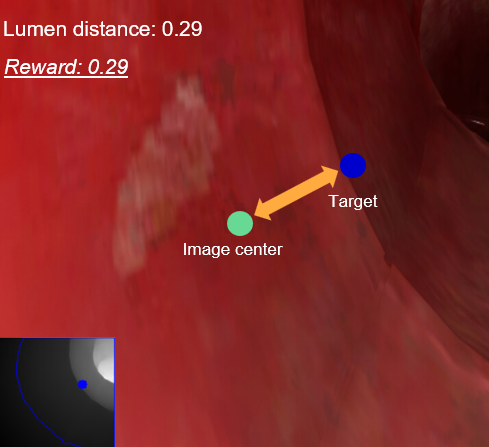}
    \caption{}
    \label{fig:1b}
  \end{subfigure}\hfill
  \begin{subfigure}[t]{0.32\columnwidth}
    \centering
    \includegraphics[width=\linewidth]{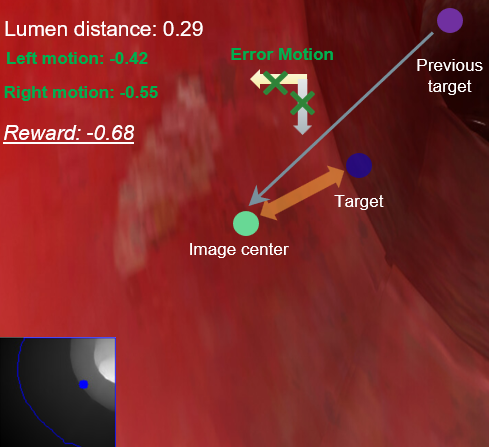}
    \caption{}
    \label{fig:1c}
  \end{subfigure}
  \caption{Reward formulation comparison. (a) and (b) share identical reward values; however, (b) is located closer to the colon wall. (c) illustrates our formulation, which explicitly accounts for previous motion errors.}
  \label{fig:1}
\end{figure}

To overcome these limitations, growing attention has focused on developing intelligent robotic systems for autonomous navigation. For example, magnetically actuated endoscopic robots have been introduced, where navigation and positioning are achieved via external magnetic field control \cite{turan2019learning}\cite{huang2021autonomous}. However, electromagnetic coil systems often occupy substantial space, limiting portability and clinical use. Another approach is vision-based navigation \cite{prendergast2020real,pore2022colonoscopy,corsi2023constrained}, which estimates lumen contours to identify the darkest region as the navigation target. Such methods adopts an end-to-end reinforcement learning framework, where raw images serve as inputs and shallow convolutional layers extract depth-related features for target localization. The objective is defined by a reward function minimizing the 2D distance between the endoscopic image center and the detected target point.

However, such methods exhibit two major limitations. First, feature extraction based on reinforcement learning frameworks such as \textit{Proximal Policy Optimization (PPO)} \cite{schulman2017proximal} is insufficient for accurate depth estimation in complex and visually dynamic environments. As a result, the identified navigation targets may be imprecise, leading to trajectory deviation. Second, during reward formulation, only the 2D lumen distance is considered while the robot’s 3D spatial position is ignored, which further amplifies this limitation. As shown in Fig.~\ref{fig:1}, although the lumen distances in Figs.~\ref{fig:1a} and \ref{fig:1b} are identical, the robot in Figs.~\ref{fig:1a} is much closer to the colon wall. In 2D models, these two distinct spatial states yield the same reward, causing inaccurate navigation feedback and a higher risk of collision.

\begin{figure*}[t]
    \centering
    \includegraphics[width=\textwidth]{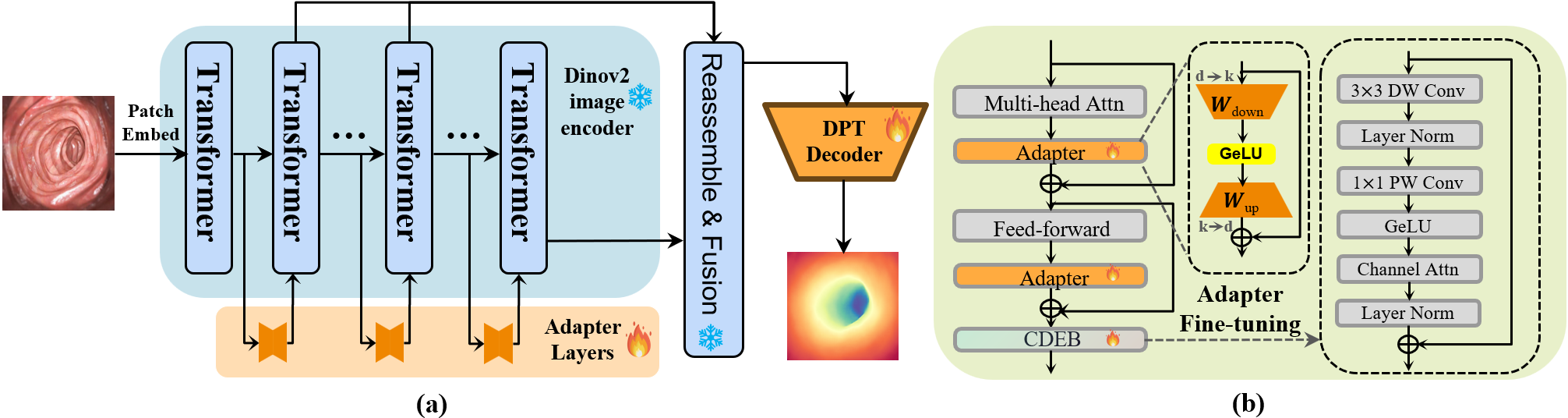}
    \caption{Overview of the proposed \textit{DepthColNet} framework. 
    (a) Fine-tuning pipeline based on \textit{Depth Anything}, where adapter layers are embedded in Transformer blocks for endoscopic domain adaptation. 
    (b) Details of the adapter and Colon Depth Enhancement Block (CDEB), while CDEB refines local geometry via depthwise separable convolution and channel attention.}

    \label{fig:3}
\end{figure*}

To address these challenges, we design a vision-based deep reinforcement learning (DRL) framework, implemented within \textit{NVIDIA Omniverse} \cite{nvidia_omniverse}. For the perception component, we adopt the foundation model \textit{Depth Anything} \cite{yang2024depth} as the visual backbone. However, since \textit{Depth Anything} is pre-trained on natural images, domain adaptation is required for endoscopic scenes. Previous approaches \cite{he2024monolot,xu2024self,yang2024self,ozyoruk2021endoslam}, constrained by the scarcity of ground-truth depth data, often relied on self-supervised transfer learning, which limited their estimation accuracy. In contrast, we employ \textit{NVIDIA Replicator} \cite{nvidia_omniverse_replicator} to synthesize realistic intraluminal images and corresponding depth maps, effectively enriching the limited real data. The model is fine-tuned on a combination of real and synthetic datasets in a supervised manner, resulting in our improved depth estimation network, \textit{DepthColNet} (Fig~\ref{fig:3}).

\textit{DepthColNet} performs monocular depth estimation and extracting navigation targets. The decision-making component then employs a DRL policy enhanced with a geometry-aware reward and penalty mechanism to achieve precise lumen following. Unlike previous 2D reward designs that solely depend on lumen distance, our approach integrates historical motion information, termed \textit{directional consistency}, to evaluate whether the robot’s movements consistently follow previously detected navigation points as shown in Fig.~\ref{fig:1c}. Deviations from the intended trajectory are penalized, enabling the agent to correct misaligned behaviors and overcome the spatial ambiguity inherent in 2D lumen-based feedback. This integration of DepthColNet with DRL ensures accurate depth perception, geometry-consistent navigation, and robust performance in complex colon environments.

\begin{figure}[t]
    \centering
    \begin{subfigure}[t]{0.48\columnwidth}
        \centering
        \includegraphics[width=\linewidth,height=2.5cm,width=2.5cm]{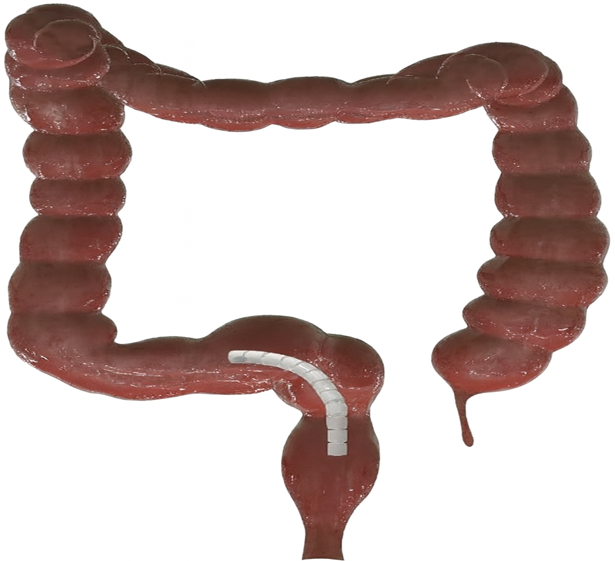}
        \caption{}
        \label{fig:2a}
    \end{subfigure}
    \begin{subfigure}[t]{0.48\columnwidth}
        \centering
        \includegraphics[width=\linewidth,height=2.5cm,width=3.5cm]{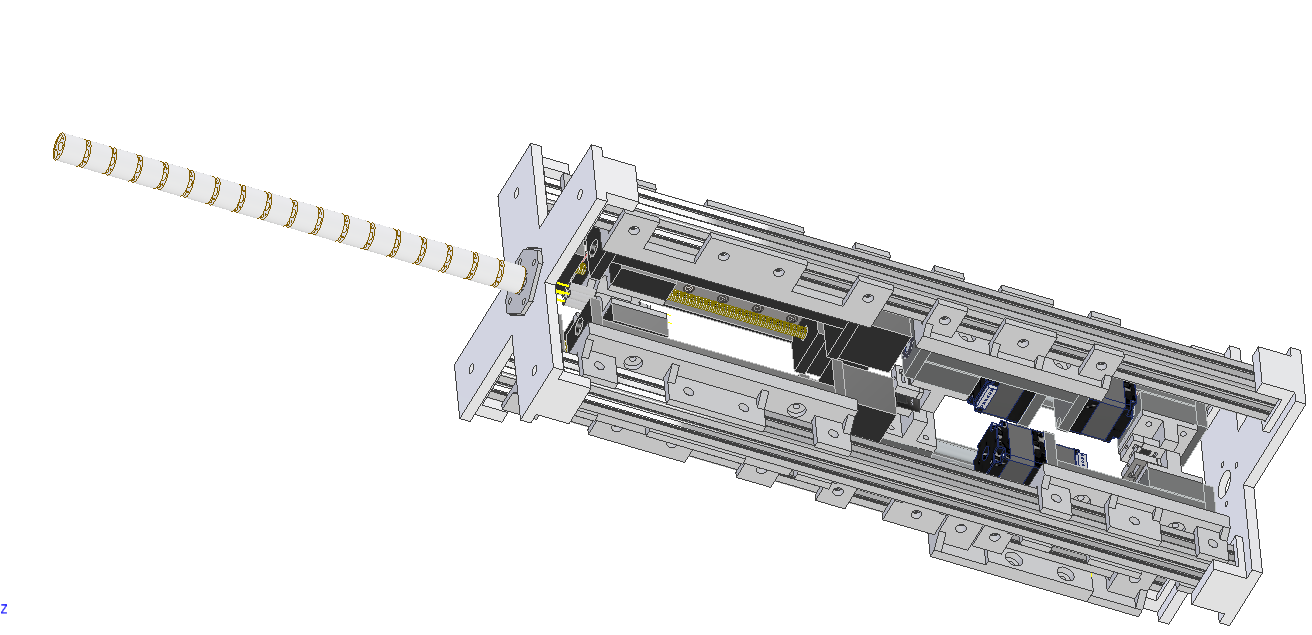}
        \caption{}
        \label{fig:2b}
    \end{subfigure}
    \caption{(a) 3D panorama of the FTL colonoscope robot, (b) FTL mechanism illustration. \vspace{-10pt}}
    \label{fig:2}
\end{figure}


In terms of robotic design, unlike traditional passive-following endoscopic robots, our system adopts an active follow-the-leader (FTL) strategy \cite{qian2025jammingsnake}, as illustrated in Fig.~\ref{fig:2}. This approach ensures that only the head section is centered within the lumen, while the trailing segments naturally align without contacting the intestinal wall, overcoming the limitations of previous passive designs that tended to press against the mucosa. This endoscopy robot is a novel continuum mechanism that achieves FTL motion through an integrated Fiber Jamming Module (FJM) \cite{qian2024effects}, with actuation provided by a rear actuator system. This study further advances this concept through systematic, simulation-based experiments, demonstrating that autonomous navigation can be achieved with a single monocular camera. This design significantly reduces hardware complexity and eliminates the need for additional actuation or magnetic systems. The main contributions of this work are summarized as follows.

\begin{enumerate}

\item We design a novel vision-based deep reinforcement learning framework, enabling fully autonomous navigation using only a monocular camera.

\item We propose \textit{DepthColNet}, a colon-specific depth estimation framework that effectively transfers the foundation model \textit{Depth Anything} to endoscopic colon environments by jointly leveraging synthetic and real data. The effectiveness of the proposed model is validated on the \textit{EndoSLAM} \cite{ozyoruk2021endoslam} dataset.


\item We conduct simulations and image synthesis using \textit{NVIDIA Omniverse} to verify the proposed system. Experimental results demonstrate that the robot autonomously navigates along the lumen centerline with superior stability and safety.
\end{enumerate}

\section{Related Work}
\textbf{Endoscopic Robot Navigation:}
Turan \textit{et al.}~\cite{turan2019learning} presented a reinforcement learning framework for magnetically actuated soft capsule endoscopes. 
Pore \textit{et al.}~\cite{pore2022colonoscopy} proposed Deep Visuomotor Control (DVC), which learns a direct mapping between endoscopic images and control signals for adaptive navigation in deformable colons. 
Corsi \textit{et al.}~\cite{corsi2023constrained} introduced formal verification that learns control directly from endoscopic images without explicit lumen detection. 
Tan \textit{et al.}~\cite{tan2025safe} developed an \textit{HI-PPO} optimization method that integrates expert knowledge through behavioral cloning, enhanced exploration, and reward-penalty adjustment to improve safety and training efficiency. 
Unlike previous methods that depend on extensive exploration in complex environments, our framework can be trained in a simple setting and still generalize effectively to more challenging scenarios.

\textbf{Colon Depth Estimation:}
Deep learning has been widely applied in the field of image processing\cite{gao2025location,gao2025multimodal,zhang2021colde,he2024monolot,jeong2024depth}.
Han \textit{et al.}~\cite{han2024depth} evaluated the zero shot \textit{Depth Anything} on endoscopic scenes, showing good generalization but limited accuracy and speed compared with domain specific models. 
Lou \textit{et al.}~\cite{lou2024surgical} fine-tuned \textit{Depth Anything} for surgical imagery, reducing blur and reflection artifacts and improving depth estimation precision. 
Ozyoruk \textit{et al.}~\cite{ozyoruk2021endoslam} proposed \textit{EndoSfMLearner}, an unsupervised framework for monocular endoscopic depth and pose estimation, also introduced the \textit{EndoSLAM} dataset for benchmarking endoscopic vision tasks.
However, these methods rely on self-supervised learning and lack spatial and motion awareness for real-time navigation. In contrast, our framework models 3D geometry and motion consistency, enabling robust lumen-centered navigation in complex colon.

\section{Method}
The objective of this work is to develop an autonomous navigation framework for endoscopic robots in complex intestinal environments. 
A high-fidelity digital twin is constructed in \textit{NVIDIA Omniverse}. 
This simulation enables large-scale data generation and policy learning in a safe, reproducible setting, reducing reliance on physical experiments. 
The proposed framework comprises three components: \textit{A.} Construction of a realistic intestinal simulation, \textit{B.} Fine-tuning of the \textit{Depth Anything} model using synthetic intraluminal images for monocular depth estimation, and \textit{C.} A reinforcement learning policy that uses depth maps as observations to achieve stable, collision-free navigation within the simulated colon.

\subsection{High-Fidelity Environment Construction}
\begin{figure}[ht]
  \centering
  \tiny
  \captionsetup[subfigure]{labelformat=empty}

  \begin{subfigure}[t]{0.3\columnwidth}
    \centering
    \includegraphics[width=\linewidth,height=2.4cm]{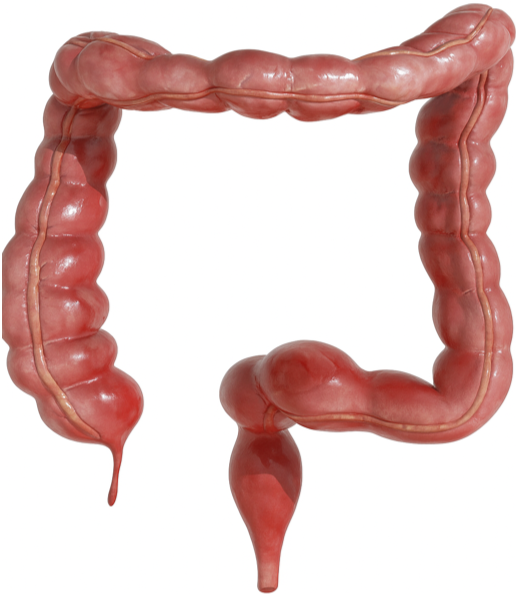}
    \caption{$C_0$}
    \label{fig:4a}
  \end{subfigure}\hfill
  \begin{subfigure}[t]{0.3\columnwidth}
    \centering
    \includegraphics[width=\linewidth]{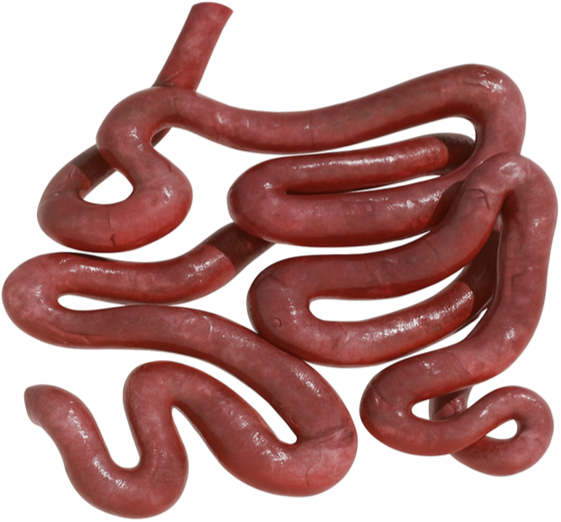}
    \caption{$C_1$}
    \label{fig:4b}
  \end{subfigure}\hfill
  \begin{subfigure}[t]{0.3\columnwidth}
    \centering
    \includegraphics[width=\linewidth,height=2.4cm]{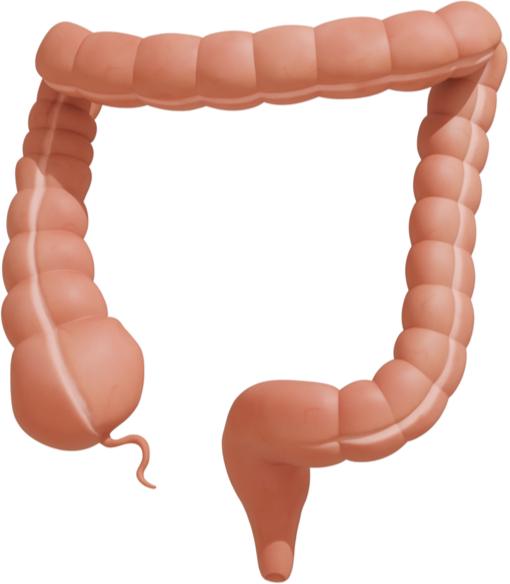}
    \caption{$C_2$}
    \label{fig:4c}
  \end{subfigure}

    \caption{$C_0$, $C_1$ share the same texture, but $C_1$ shows a more complex path. $C_0$, $C_2$ have similar paths but differ in texture.}
  \label{fig:4}
\end{figure}

A high-fidelity intestinal simulation environment is built in \textit{NVIDIA Omniverse} to replicate realistic anatomy, illumination, and camera motion. 
This virtual platform supports large-scale data generation and safe reinforcement learning without physical risks. 
As shown in Fig.~\ref{fig:4}, three colon models are constructed, including two simplified and one anatomically complex. 
The simplified models differ in surface textures to assess the robustness of depth estimation under visual noise, while all models include realistic physical properties such as friction and collision response, ensuring both perceptual diversity and dynamic fidelity for reliable policy learning.

\subsection{Monocular Depth Estimation Network}

\textit{Depth Anything} is a foundation model for natural image depth estimation that employs DINOv2~\cite{oquab2024dinov2} as the encoder and a Dense Prediction Transformer (DPT)~\cite{ranftl2021vision} as the decoder. 
As illustrated in Fig.~\ref{fig:3}(a), we fine-tune the model using a hybrid dataset of real and synthetic endoscopic images to achieve domain adaptation for colonoscopy. 
To address the domain gap in endoscopic scenes, we design \textit{Adapters} specialized for colon imagery, as shown in Fig.~\ref{fig:3}(b). 
The proposed \textit{DepthColNet} introduces two core components: adapter layers for efficient backbone adaptation and a Colon Depth Enhancement Block (CDEB) that refines discriminative geometric cues critical for accurate depth perception in colon environments.

\subsubsection{Adapter-Based Backbone Fine-Tuning}
The concept of \textit{adapters} \cite{han2024parameter} was originally introduced for multi-domain visual classification to achieve efficient parameter transfer across different visual domains by incorporating lightweight, trainable modules into pretrained networks. In Transformer-based architectures, adapters are typically inserted after the self-attention and feed-forward layers, enabling new tasks to be learned with minimal additional parameters while retaining the generalization capability of the pretrained backbone.  

Each adapter consists of a down-projection layer $W_{\text{down}} \in \mathbb{R}^{d\times k}$, a $\mathrm{GeLU}$ activation ($\sigma$), and an up-projection layer $W_{\text{up}} \in \mathbb{R}^{k\times d}$, connected through a residual path:
\begin{equation}
\mathbf{X}' = \mathbf{X} + W_{\text{up}}\,\sigma(W_{\text{down}}\mathbf{X}),
\end{equation}
where $d$ is the feature dimension and $k \ll d$ is the adapter bottleneck size. During fine-tuning, only $W_{\text{up}}$ and $W_{\text{down}}$ are updated, while all pretrained parameters of \textit{Depth Anything} remain frozen. This design effectively adapts the model to the domain shift between natural and endoscopic images, capturing colon-specific depth cues while preventing overfitting.  

Unlike \textit{LoRA} \cite{hu2022lora} (Low-Rank Adaptation), which applies low-rank linear updates to attention weights ($\Delta W = BA$), the adapter introduces nonlinear residual mappings that better capture spatially varying depth cues in endoscopic scenes, ensuring consistent and robust geometric representation.

\subsubsection{Colon Depth Enhancement Block (CDEB)}
Transformers are powerful in capturing global dependencies but tend to lose fine-grained structural cues in endoscopic images, where subtle geometric variations and depth discontinuities are critical. To compensate for this limitation, we design a Colon Depth Enhancement Block (CDEB) that reinforces local geometric representation while maintaining global semantic consistency, as illustrated in Fig.~\ref{fig:3}b. CDEB comprises a $3\times3$ depthwise and a $1\times1$ pointwise convolution (DW+PW) followed by a channel attention module. The DW+PW structure efficiently extracts intestinal contours and depth gradients with low complexity, while channel attention adaptively emphasizes depth-relevant features. Enhanced features are fused through a residual connection to maintain stability and depth continuity. To balance efficiency and accuracy, CDEBs are inserted after the 3rd, 5th, 8th, and 11th Transformer layers whose features are fed into the DPT decoder.


\begin{figure}[ht]
  \centering
  \includegraphics[width=0.8\columnwidth,height=5.5cm]{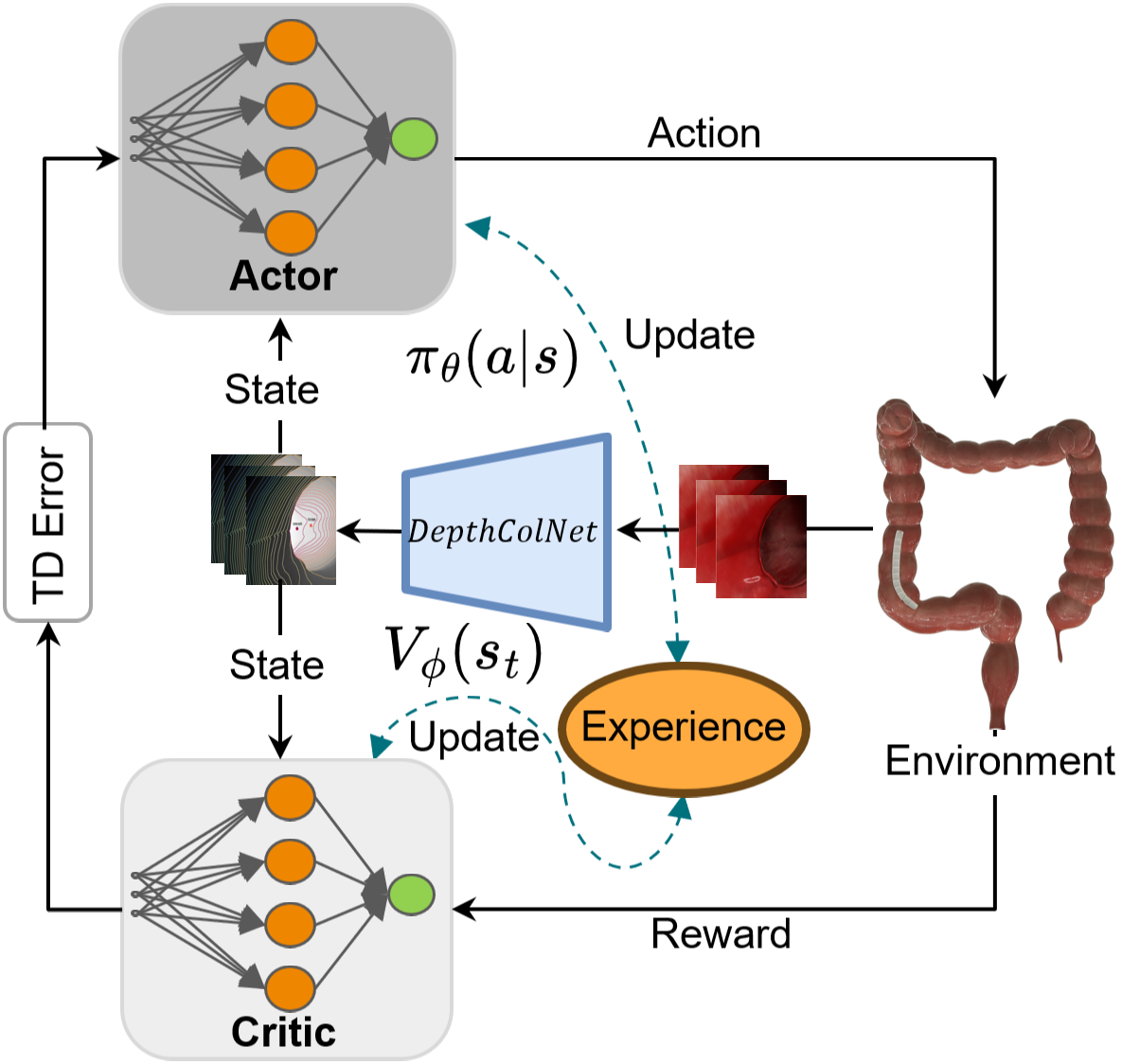}
  \caption{The overview of our proposed vision-based deep reinforcement learning navigation framework.}
  \label{fig:rl_frame}
\end{figure}

\subsection{Reinforcement Learning for Autonomous Navigation}
\subsubsection{Deep Reinforcement Learning}
The autonomous colon navigation problem is formulated as a \textit{Markov Decision Process (MDP)}, defined by the tuple $(\mathcal{S}, \mathcal{A}, \mathcal{R}, \mathcal{P}, \gamma, H)$, 
where $\mathcal{S}$, $\mathcal{A}$, and $\mathcal{R}$ represent the state, action, and reward spaces, respectively; 
$\mathcal{P}$ denotes the transition probability distribution; $\gamma \in [0,1]$ is the discount factor; and $H$ is the episode horizon. 
At each time step $t$, the environment produces an observation $s_t \in \mathcal{S}$, and the agent executes an action $a_t \in \mathcal{A}$ according to a stochastic policy $a_t \sim \pi_\theta(a_t|s_t)$. 
After execution, the agent receives a scalar reward $r_t = \mathcal{R}(s_t,a_t)$ and transitions to $s_{t+1} \sim \mathcal{P}(s_{t+1}|s_t,a_t)$. 
The objective is to find an optimal policy that maximizes the cumulative discounted reward:
\begin{equation}
J(\pi_\theta) = \mathbb{E}_{\tau \sim \pi_\theta}\!\left[\sum_{t=0}^{H}\gamma^t r_t\right].
\end{equation}

As shown in Fig.~\ref{fig:rl_frame}, we employ an \textit{Actor--Critic} \cite{bahdanau2016actor} framework, where the actor network parameterizes the policy $\pi_\theta(a|s)$, and the critic network estimates the value function $V_\phi(s)$ to evaluate expected returns. 
The critic minimizes the temporal-difference (TD) error $\mathcal{L}_V = \mathbb{E}[(V_\phi(s_t) - R_t)^2]$, 
where $R_t$ is the discounted return and the advantage function is computed as $A_t = R_t - V_\phi(s_t)$.
To stabilize training, we adopt the \textit{PPO} \cite{schulman2017proximal} algorithm, which constrains large policy updates by clipping the probability ratio 
$r_t(\theta) = \pi_\theta(a_t|s_t)/\pi_{\theta_{\text{old}}}(a_t|s_t)$ within $[1-\epsilon, 1+\epsilon]$. 

\begin{figure}[ht]
    \centering
    \begin{subfigure}[t]{0.48\columnwidth}
        \centering
        \includegraphics[height=2.2cm,width=\linewidth]{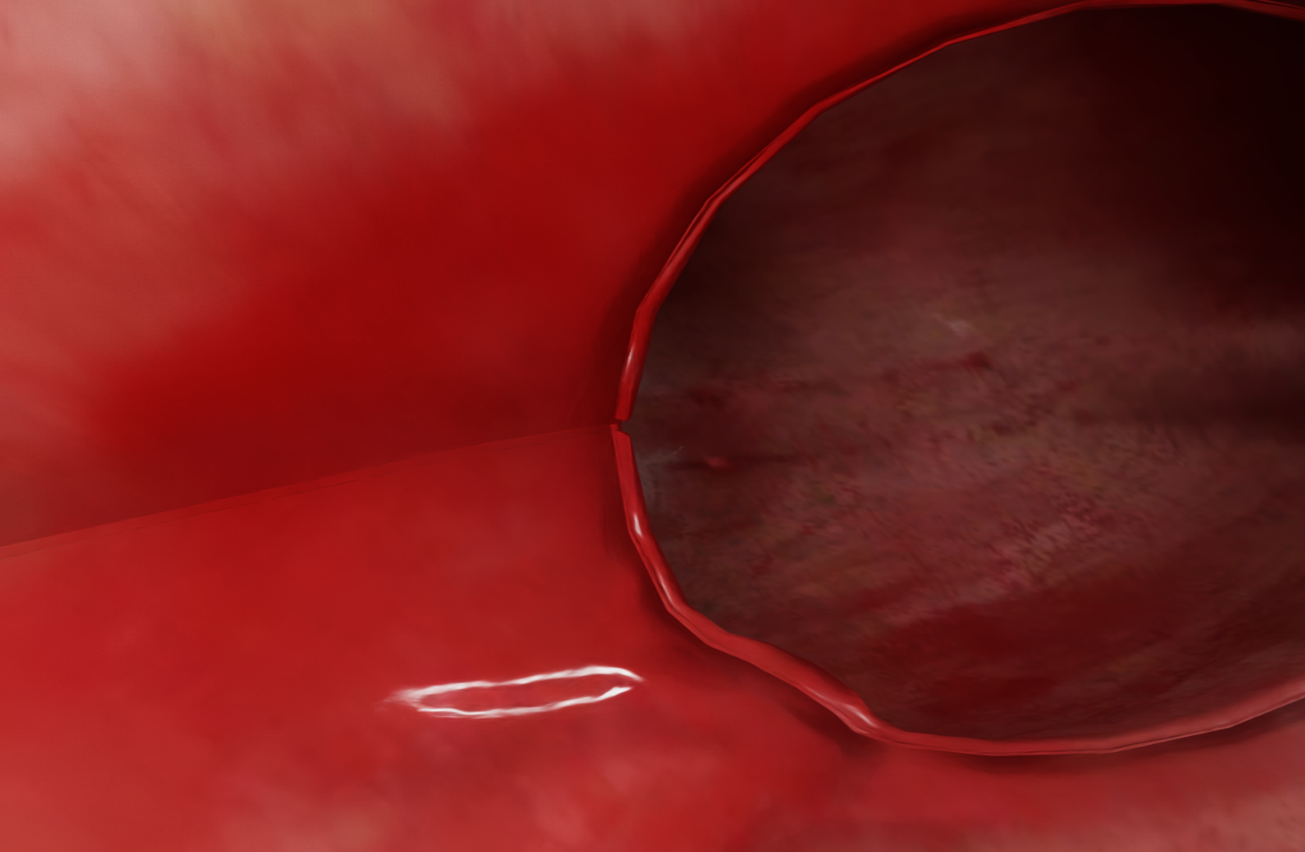}
        \caption{Raw Image}
        \label{fig:tara}
    \end{subfigure}
    \hfill
    \begin{subfigure}[t]{0.48\columnwidth}
        \centering
        \includegraphics[height=2.2cm,width=\linewidth]{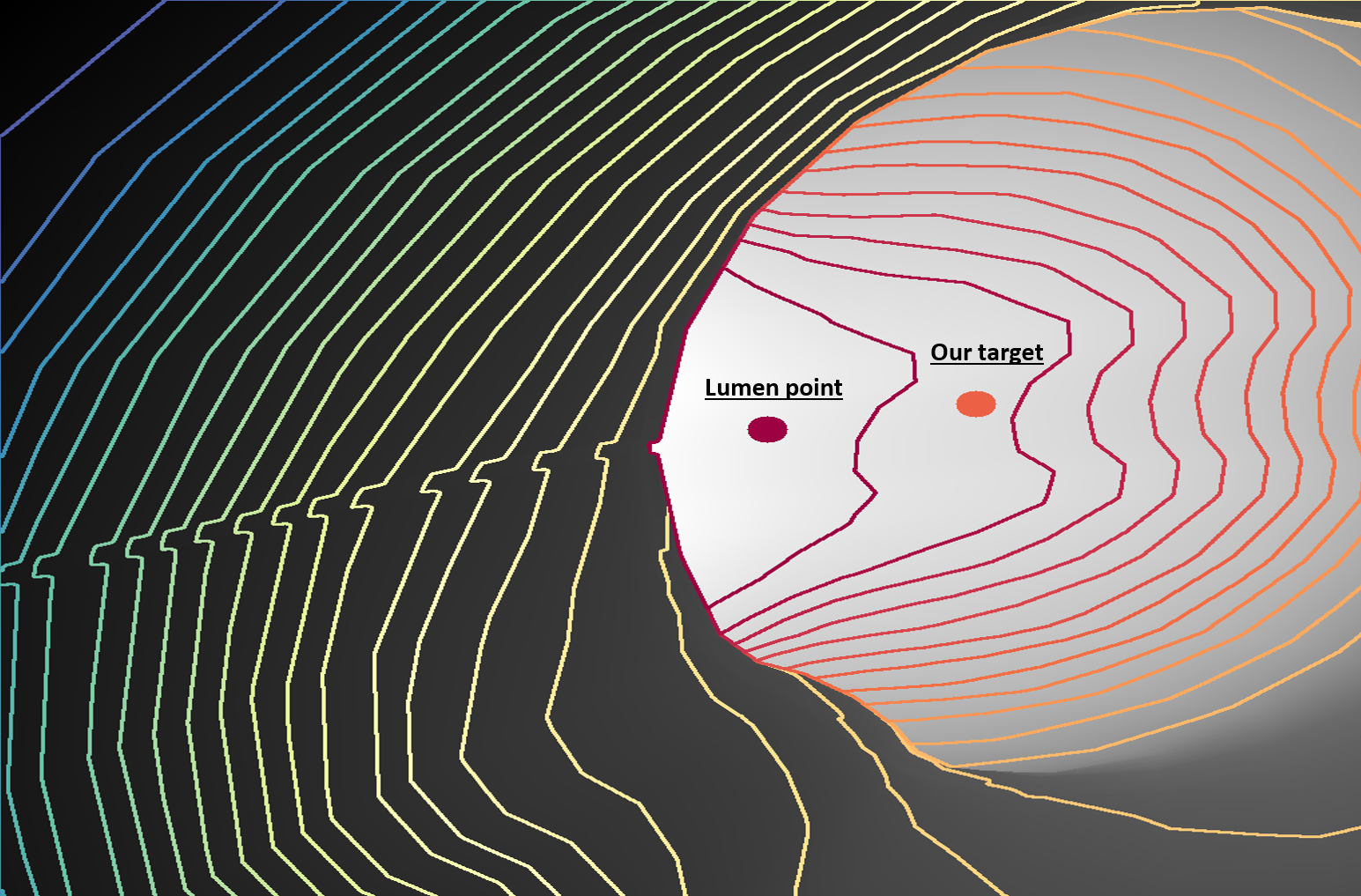}
        \caption{Depth Contour Map}
        \label{fig:tarb}
    \end{subfigure}
    \caption{Illustration of the navigation point extraction process, depth contour visualization with lumen and our target points.}
    \label{fig:tar}
\end{figure}

\subsubsection{Observation Space}
At each time step, the agent receives an RGB image from the endoscopic camera. 
This image is processed by our \textit{DepthColNet} model to produce a depth map. 
In previous approaches \cite{prendergast2020real,pore2022colonoscopy,corsi2023constrained}, navigation relied on the lumen point, but we observed that this often leads to premature turning and wall collisions in curved regions. 
As shown in Fig.~\ref{fig:tar}, following the lumen point would cause the robot to steer into the left wall. In contrast, we uniformly divide the depth values into 20 quantile levels and, through evaluation, find that using the centroid of the 8th-level depth contour as the navigation point yields the most stable performance. Given the target $(t_x,t_y)$ and the image center $(c_x,c_y)$, the observation is encoded as a 4D normalized feature:
\begin{equation}
\mathbf{o}_t = [x_{\text{norm}}, y_{\text{norm}}, \Delta x_{\text{norm}}, \Delta y_{\text{norm}}],
\end{equation}
where $(x_{\text{norm}}, y_{\text{norm}})$ are globally normalized coordinates within $[-1,1]$, 
and $(\Delta x_{\text{norm}}, \Delta y_{\text{norm}})$ denote the normalized relative displacement, computed as 
$(\Delta x_{\text{norm}}, \Delta y_{\text{norm}}) = (2(t_x-c_x)/W - 1,\; 2(t_y-c_y)/H - 1)$, 
which encodes the offset from the image center to the navigation target in a scale-invariant form, providing stable spatial cues for navigation.

\subsubsection{Action Space}
The endoscope is modeled based on a follow-the-leader (FTL) design, in which the trailing body passively follows the trajectory traced by the tip. Under this design, safe navigation can be ensured by controlling tip motion alone, provided the tip avoids collisions with the colonic wall. Therefore, we allow only tip-level motions, including up–down and left–right directional adjustments as well as forward–backward axial translation.
The agent operates in a continuous 3D action space designed to control the motion of the endoscope tip during navigation. 
Each action is represented as a vector $\mathbf{a}_t = (\Delta x, \Delta y, \Delta z) \in [-1,1]^3$, 
corresponding to normalized pitch, yaw, and forward–backward translation along the endoscope’s local axes. 
The angular components $(\Delta x, \Delta y)$ are scaled to produce smooth directional adjustments, enabling the agent to align the viewing direction with the lumen orientation, 
while $\Delta z$ controls the translational motion along the insertion axis.  

A constant forward velocity of $1\,\mathrm{mm/s}$ is applied only when the lumen region is visible in the predicted depth map—specifically, when the ratio between the target–center distance and the diagonal–center distance is below 35\%; 
otherwise, the translational component is suppressed to prevent wall collisions.  
This continuous formulation allows fine-grained steering and adaptive motion gating, ensuring stable, collision-free, and visibility-aware autonomous navigation within the simulated colon environment.

\subsubsection{Reward Function}

The reinforcement learning agent is optimized under a composite reward function that integrates spatial alignment, motion stability, and navigation efficiency. 
At each timestep, the total reward is defined as:
\begin{equation}
r_t =
\,r_{\text{dis}}(t)
+ \,r_{\text{dir}}(t)
+ \,r_{\text{succ}}(t)
+ r_{\text{step}}
+ r_{\text{penalty}}(t),
\end{equation}
where each term serves a distinct functional purpose.

\textbf{Distance Reward:}  
Given an image \( W \times H \), the target centroid is located at \( (c_x, c_y) \). 
Its normalized coordinates relative to the image center \( (\tfrac{W}{2}, \tfrac{H}{2}) \) are computed as:
$u_t = \frac{c_x - \tfrac{1}{2}W}{\tfrac{1}{2}W}, \quad
v_t = \frac{c_y - \tfrac{1}{2}H}{\tfrac{1}{2}H}$.
The distance reward is defined by:
\begin{equation}
r_{\text{dis}}(t) = 1 - \sqrt{u_t^2 + v_t^2 + \varepsilon},
\end{equation}
where \( \varepsilon = 10^{-6} \) ensures numerical stability. 
$u_t$ and $v_t$ denote the normalized horizontal and vertical offsets between the target and the image center. This reward encourages the navigation target to remain near the image center, guiding the endoscope to align its viewing axis with the lumen path.

\textbf{Direction Consistency Reward:}  
To ensure that the control action aligns with the target’s displacement direction, a directional consistency term is introduced:
\begin{equation}
r_{\text{dir}}(t)
= \frac{a_{\mathrm{lr}}u_t + a_{\mathrm{ud}}v_t}
{\sqrt{u_t^2 + v_t^2 + \varepsilon}},
\end{equation}
where \( a_{\mathrm{lr}} \) and \( a_{\mathrm{ud}} \) represent the agent’s lateral and vertical control signals.  
This term measures the cosine similarity between the control vector and the target offset, rewarding movements that steer the endoscope tip toward the lumen center and penalizing those that deviate from it.

\textbf{Success Reward:}  
A high success reward is granted when the target lies within a tolerance \( \tau \) around the image center:
\begin{equation}
r_{\text{succ}}(t) =
\begin{cases}
300, & \text{if } |u_t| < \tau \text{ and } |v_t| < \tau,\\[4pt]
0, & \text{otherwise.}
\end{cases}
\end{equation}
This component provides a strong positive signal when alignment is achieved, reinforcing precise positioning behavior.

\textbf{Step Penalty:}  
A small per-step penalty discourages excessively long trajectories:
\begin{equation}
r_{\text{step}} =
\begin{cases}
0, & \text{if success},\\[4pt]
-0.015, & \text{otherwise.}
\end{cases}
\end{equation}
Motivating the agent to complete navigation efficiently.

\textbf{Stability Penalty:}  
To prevent oscillatory or unstable actions, an additional penalty is applied if more than five of the last \( N \) rewards are non-positive:
\begin{equation}
r_{\text{penalty}}(t) =
\begin{cases}
-0.5, & \text{if } \sum_{i=t-N}^{t} [r_i \le 0] \ge 5,\\[4pt]
0, & \text{otherwise.}
\end{cases}
\end{equation}
Here, \( N=10 \) denotes the temporal window size and \( r_i \) the past rewards.  
This penalty discourages repetitive unproductive actions and promotes smooth, stable movements.

\section{Experiments}
\subsection{Implementation Details and Datasets}
Training and evaluation were performed on three colon models constructed in Blender, following the anatomical references in \cite{Sundberg2020_small_large_intestine,HumanIntestines_Fab}, as shown in Fig.~\ref{fig:4}. Among them, models \(\mathrm{C_0}\) and \(\mathrm{C_2}\) share the same geometry but differ in texture, whereas \(\mathrm{C_1}\) features a more complex geometry while retaining the texture of \(\mathrm{C_0}\). The depth model was fine-tuned on the \textit{ColonDepth} dataset and our synthetic data generated from \(\mathrm{C_0}\), and validated using the \textit{EndoSLAM} dataset. The colon environment \(\mathrm{C_0}\) was used for training, while \(\mathrm{C_1}\) and \(\mathrm{C_2}\) were used to evaluate the effectiveness and generalization capability of our reinforcement learning navigation algorithm.

\textbf{Real data:}
Following prior studies \cite{lou2024surgical,han2024depth
}, for endoscopic depth estimation, two datasets are employed for training and evaluation. 
The \textit{ColonDepth}~\cite{rau2019implicit} dataset contains 16{,}016 colon RGB images with corresponding depth maps for training. 
For testing, we adopt the \textit{EndoSLAM}~\cite{ozyoruk2021endoslam} dataset, which includes 21{,}887 colon image–depth pairs.

\begin{figure}[h]
  \centering
  \begin{subfigure}[t]{0.32\columnwidth}
    \centering
    \includegraphics[width=\linewidth]{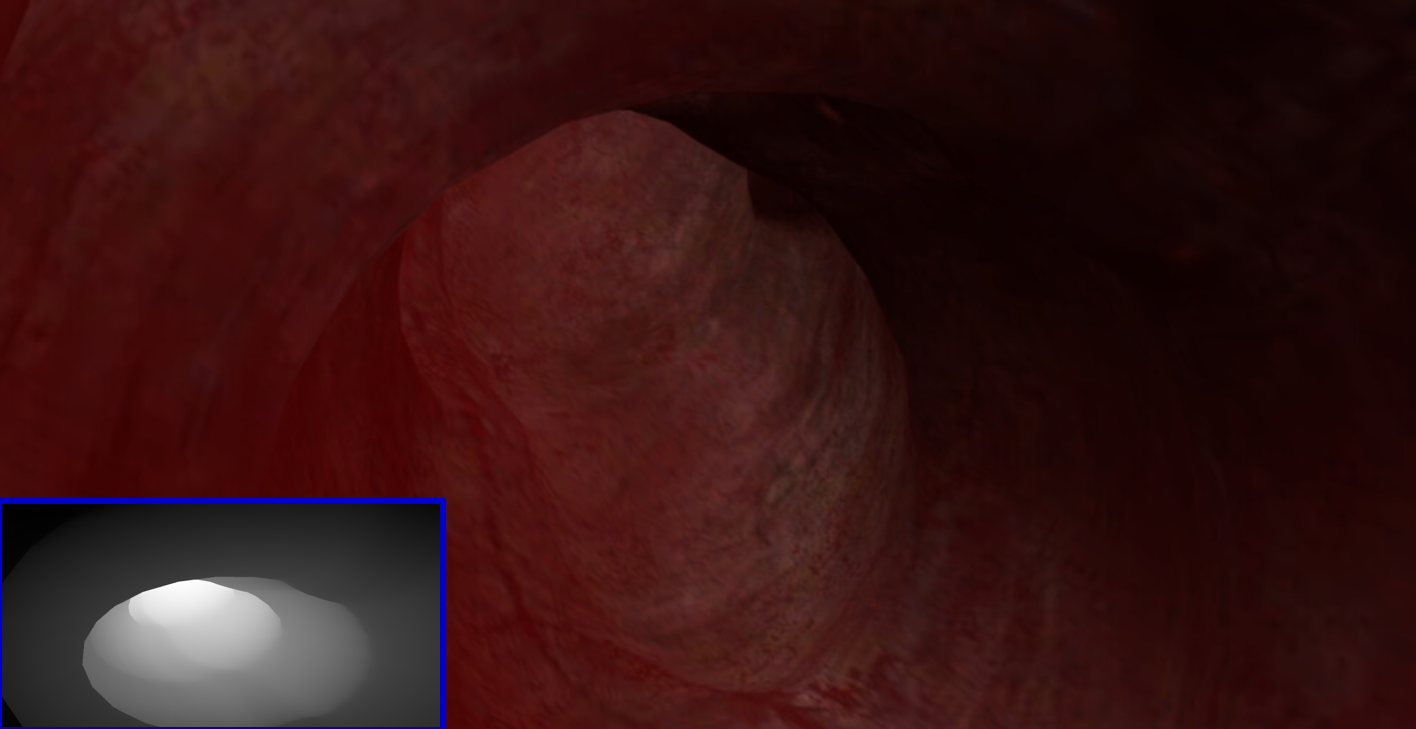}\\[2pt]
    \caption{Low Illumination.}
    \label{fig:5a}
  \end{subfigure}\hfill
  \begin{subfigure}[t]{0.32\columnwidth}
    \centering
    \includegraphics[width=\linewidth]{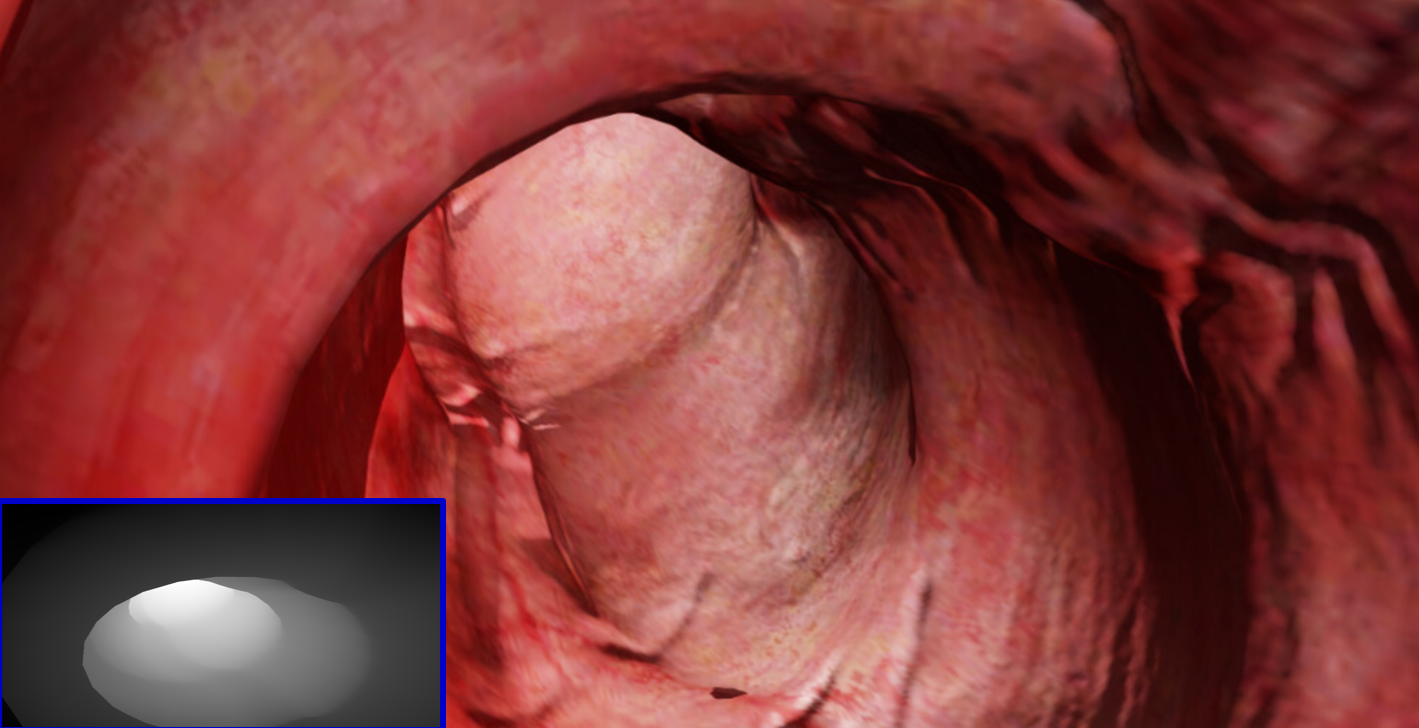}\\[2pt]
    \caption{High Illumination.}
    \label{fig:5b}
  \end{subfigure}\hfill
  \begin{subfigure}[t]{0.32\columnwidth}
    \centering
    \includegraphics[width=\linewidth]{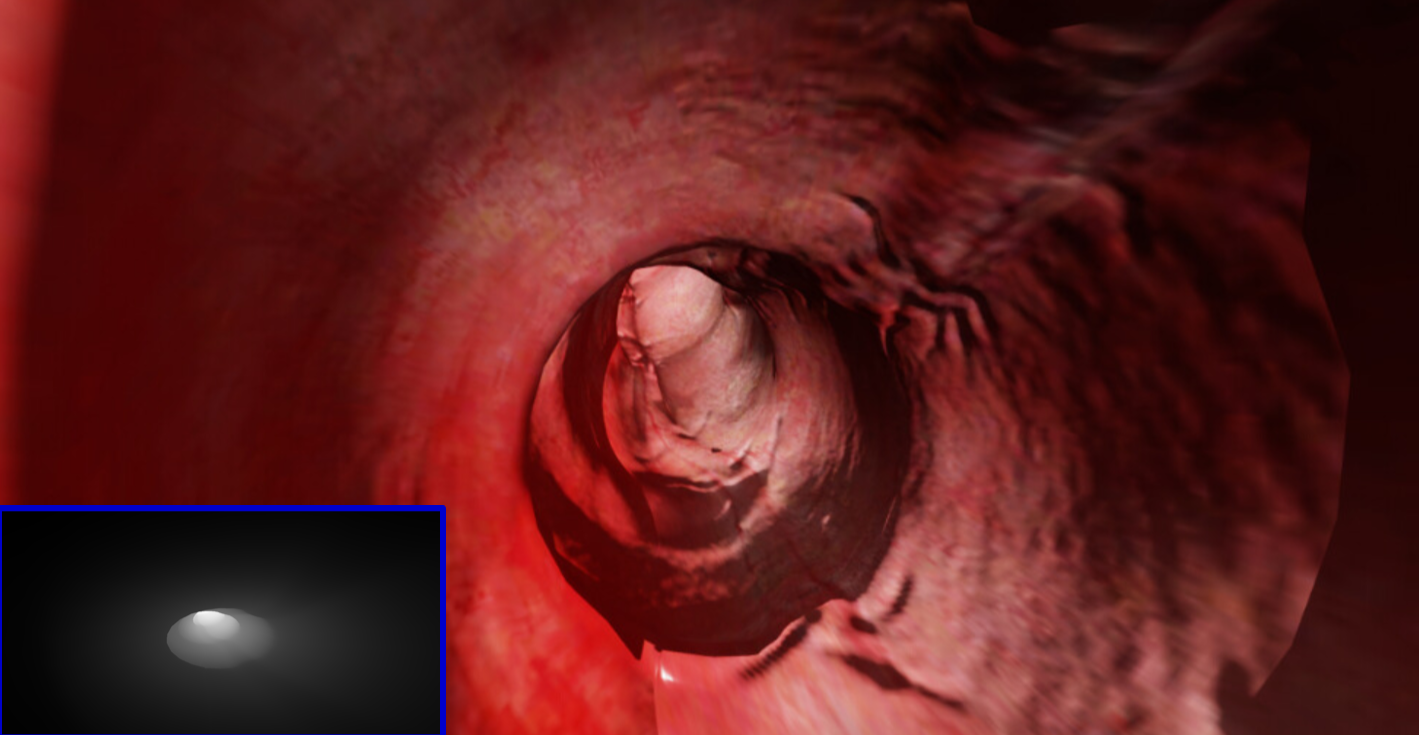}\\[2pt]
    \caption{Adjusted FOV.}
    \label{fig:5c}
  \end{subfigure}

  \caption{Example of synthetic data with corresponding depth ground truth. (a), (b) share the same FOV but differ in illumination, while (c) changes the FOV of (b).}
  \label{fig:5}
\end{figure}

\textbf{Synthetic data:}
Within the Omniverse simulation environment, we employ \textit{NVIDIA Replicator} \cite{nvidia_omniverse_replicator} to generate 5,068 synthetic intraluminal images and their corresponding depth maps, as illustrated in Fig.~\ref{fig:5}. The dataset is constructed by randomizing the robot’s initial positions, and lighting intensities within the colon model ($C_0$) to enhance visual diversity. This combination of real and simulated data effectively bridges the domain gap, enabling the fine-tuned model to achieve robust monocular depth estimation in challenging endoscopic scenes.

For the fine-tuning of our depth estimation model, the framework is implemented using \textit{PyTorch}. 
The \textit{AdamW} optimizer is employed with an initial learning rate of \(3\times10^{-5}\). 
The warm-up step is set to 5000, and the scaling factor for the loss term \(L_p\) is \(\alpha = 0.85\). 
A batch size of 8 is used, and the model is trained for 50 epochs in total. 

For the DRL experiments, all training and evaluation were conducted using \textit{NVIDIA Isaac Sim v5.0.0} and \textit{NVIDIA Isaac Lab v2.2}. 
The actor-critic networks shared a lightweight MLP backbone with hidden dimensions of [128, 64], employing \textit{ELU} activations and \textit{D2RL} connections to enhance feature reuse and stabilize training. 
Training was performed in \textit{RL-Games} framework, using \(\gamma = 0.99\), \(\tau = 0.95\), a learning rate of \(3\times10^{-4}\), a batch size of 16 for approximately 1.5 million steps until convergence. 
Three complementary metrics-geometric alignment, navigation safety, and motion smoothness were adopted for quantitative evaluation, as summarized in Table~\ref{tab:metrics}. 
All experiments were executed on a workstation equipped with an \textit{NVIDIA RTX 5090 GPU}.


\begin{table*}[ht]
\centering
\caption{Navigation metrics used for quantitative assessment.}
\renewcommand{\arraystretch}{1.22}
\resizebox{\textwidth}{!}{%
\begin{tabular}{l l p{11.8cm} c}
\toprule
\textbf{Metric} & \textbf{Formula} & \textbf{Definition / Description} & \textbf{Better} \\
\midrule

\textbf{Geometry-aware Lumen Distance ($D_{\text{geo}}$)}
&
$D_{\text{geo}} = \tfrac{1}{N}\!\sum_{i=1}^{N}\!\left(\tfrac{d_i}{d_{\max}}\right)\!\left(1+0.5\tfrac{\|p_i-\hat{p}_i\|_2}{r_i}\right)$
&
Evaluates 3D lumen alignment by combining 2D centering and geometric consistency.
$d_i$ denotes the distance between the image center and the target point, normalized by $d_{\max}$.
$|p_i-\hat{p}_i|_2$ measures the spatial deviation from the ideal centerline projection, normalized by the local colon radius $r_i$, which ensures discrimination between targets with similar 2D distances but different 3D positions.
& $\downarrow$ \\

\textbf{Navigation Safety Score ($S_{\text{nav}}$)}
&
$S_{\text{nav}} = 1 - \!\Big(0.6\tfrac{N_{\mathrm{col}}}{N_{\mathrm{step}}} + 0.4\tfrac{D_{\mathrm{p}}}{L_c}\Big)$
&
Evaluates navigation safety and efficiency based on collision and path performance.
$N_{\mathrm{col}}/N_{\mathrm{step}}$ represents the collision rate,
$D_{\mathrm{p}}$ denotes the actual path length of the robot,
and $L_c$ is the length of the colon centerline.
& $\uparrow$ \\

\textbf{Jerk Index ($J$)}
&
$J = \tfrac{1}{T}\!\int_{0}^{T}\!\big\lVert \tfrac{d^{3}x}{dt^{3}} \big\rVert dt$
&
Measures motion smoothness, where smaller values indicate smoother and more stable control behavior.
& $\downarrow$ \\

\bottomrule
\end{tabular}
}
\label{tab:metrics}
\end{table*}


\vspace{-5.5pt}

\vspace{-8pt}

\begin{figure}[ht]
  \centering
  \begin{subfigure}[t]{0.32\columnwidth}
    \centering
    \includegraphics[width=\linewidth,height=1.7cm]{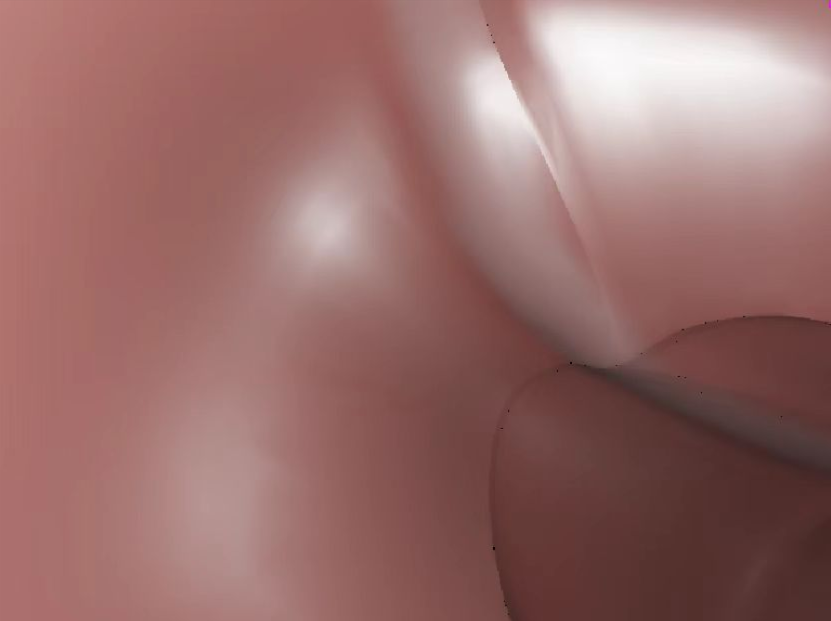}
    \caption{Raw Image}
    \label{fig:va}
  \end{subfigure}\hfill
  \begin{subfigure}[t]{0.32\columnwidth}
    \centering
    \includegraphics[width=\linewidth,height=1.7cm]{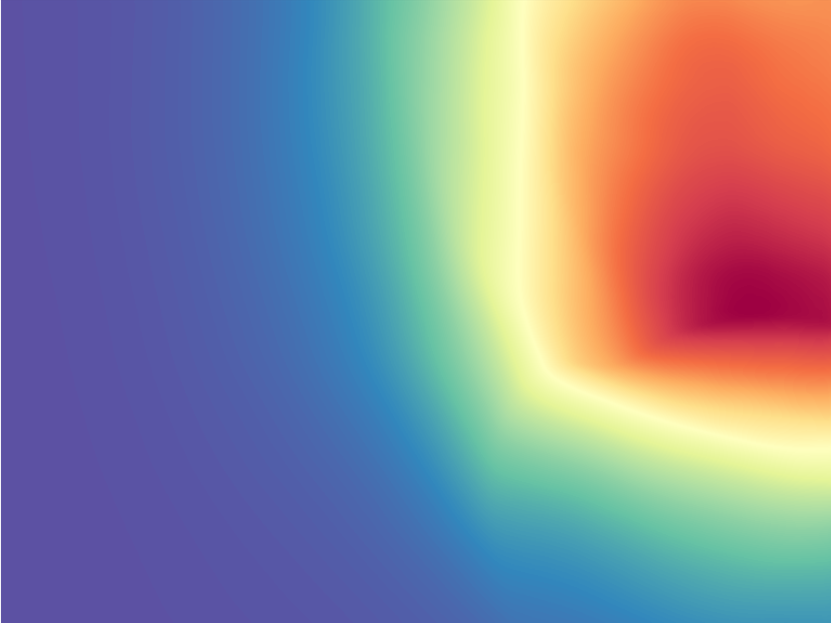}
    \caption{\textit{Depth Anything}}
    \label{fig:vb}
  \end{subfigure}\hfill
  \begin{subfigure}[t]{0.32\columnwidth}
    \centering
    \includegraphics[width=\linewidth,height=1.7cm]{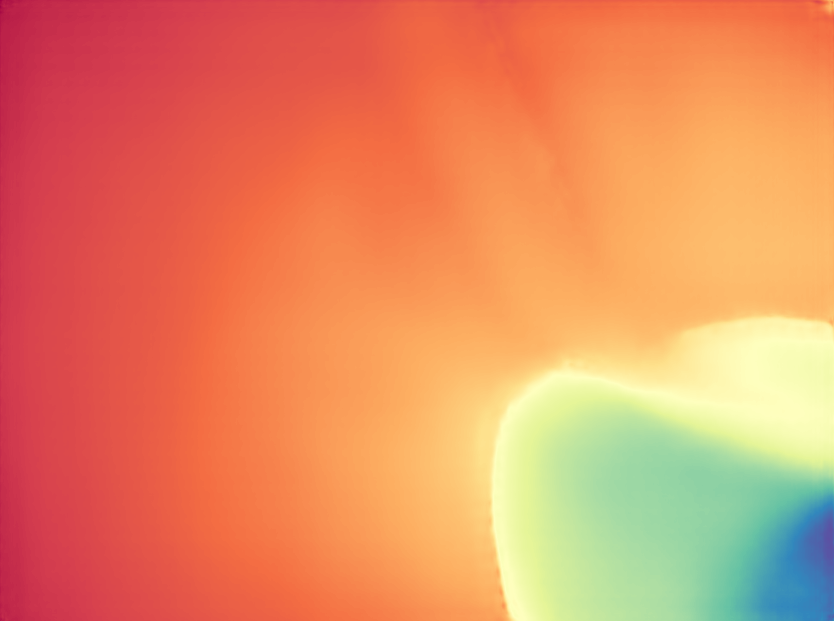}
    \caption{Ours}
    \label{fig:vc}
  \end{subfigure}
  %
  \caption{Depth visualization comparison between ours and \textit{Depth Anything}. Cooler colors (blue) indicate greater depth.}

  \label{fig:v}
\end{figure}

\begin{table}[ht]
\centering
\caption{Comparison of depth estimation methods.}
\renewcommand{\arraystretch}{1.2}
\resizebox{\columnwidth}{!}{%
\begin{tabular}{l l c c c}
\toprule
\textbf{} & \textbf{Methods} & \textbf{ZeroShot} & \textbf{\shortstack{ColonDepth \\ (\& Synthetic)}} & \textbf{ColonDepth} \\
\midrule

\multirow{3}{*}{\textbf{Abs. Rel. ($\downarrow$)}} 
 & Endo-SfMLearner \cite{ozyoruk2021endoslam} & \textbf{0.551} & 0.512 & 0.551 \\
 & Depth Anything \cite{yang2024depth} & 0.616 & 0.284 & 0.306 \\
 & DepthColNet (Ours) & \textemdash & \textbf{0.245} & \textbf{0.262} \\
\midrule

\multirow{3}{*}{\textbf{$\boldsymbol{\delta_{1}(\mkern-2mu\uparrow\mkern-2mu)}$}} 
 & Endo-SfMLearner \cite{ozyoruk2021endoslam} & \textbf{0.354} & 0.376 & 0.354 \\
 & Depth Anything \cite{yang2024depth} & 0.285 & 0.648 & 0.622 \\
 & DepthColNet (Ours) & \textemdash & \textbf{0.677} & \textbf{0.654} \\
\bottomrule
\end{tabular}
}
\label{tab:depth_comparison}
\end{table}

\subsection{Quantitative Evaluation on Depth Estimation}

We compare our method with two representative approaches under three settings, as shown in Table~\ref{tab:depth_comparison} and Fig.~\ref{fig:v}. 
In the Zero-Shot setting, models are directly evaluated on \textit{EndoSLAM} without fine-tuning~\cite{han2024depth}. 
In the \textit{ColonDepth} and \textit{ColonDepth+Synthetic} settings, models are fine-tuned on real images only or jointly with synthetic colon data from Omniverse~\cite{lou2024surgical}. 
\textit{Endo-SfMLearner}~\cite{ozyoruk2021endoslam} serves as the baseline trained on monocular endoscopic videos, while \textit{Depth Anything}~\cite{yang2024depth} represents a large-scale foundation model for natural images. For evaluation, we adopt standard depth estimation metrics. Absolute Relative Error ($\text{Abs. Rel.}$) measures the average relative deviation between the predicted and ground truth depth values, while $\delta_{1}$ accuracy quantifies the proportion of depth estimates that fall within a predefined tolerance of the ground truth.
Our fine-tuned model reduces $\text{Abs. Rel.}$ by $4.4\%$ and improves $\delta_{1}$ by $3.2\%$ compared with \textit{Depth Anything}, showing better robustness to specular reflections and texture variations in colonoscopy scenes. 
Adding synthetic data further decreases $\text{Abs. Rel.}$ by $1.7\%$ and increases $\delta_{1}$ by $2.3\%$. 
Since our framework introduces task-specific modules, zero-shot comparisons are not directly applicable.


\subsection{Quantitative Evaluation on Autonomous Navigation}
We compare our method with three representative baselines on the $C_1$ colon environment, which has more frequent and complex bends that better highlight the navigation performance of different methods.
As shown in Table~\ref{tab:navi_comparison}, our approach achieves the best performance across all metrics, outperforming the second-best method by $0.22$, $0.21$, and $0.67$ on $S_{\text{geo}}$, $S_{\text{nav}}$, and $J$, respectively. 
The proposed $S_{\text{geo}}$ metric accounts for the robot’s 3D spatial position when computing the lumen distance, effectively reflecting geometric alignment with the colon centerline. 
In contrast, competing methods tend to deviate from the centerline even when the 2D lumen distances appear similar, increasing the risk of navigation drift. 
The $S_{\text{nav}}$ score evaluates both collision rate and trajectory optimality, where our method demonstrates superior safety and efficiency under complex anatomical conditions. 
Furthermore, the lowest Jerk Index indicates that our approach achieves the smoothest and most stable motion, ensuring precise and safe lumen-centered navigation.

\begin{table}[ht]
\centering
\tiny  
\caption{Comparison of navigation performance among different methods in the colon environment $C_1$.}
\renewcommand{\arraystretch}{1.2}
\resizebox{\columnwidth}{!}{%
\begin{tabular}{lccc}
\toprule
\textbf{Method} & $S_{\text{geo}}$ ($\downarrow$) & $S_{\text{nav}}$ ($\uparrow$) & $J$ ($\downarrow$) \\
\midrule
\textit{L-PPO}~\cite{pore2022colonoscopy} & 1.36 ± 0.12 & 0.53 ± 0.01 & 3.24 ± 0.05 \\
\textit{C-PPO}~\cite{corsi2023constrained} & 1.04 ± 0.33 & 0.59 ± 0.04 & 2.76 ± 0.13 \\
\textit{HI-PPO}~\cite{tan2025safe} & 0.83 ± 0.02 & 0.73 ± 0.06 & 2.39 ± 0.07 \\
\midrule
Ours & \textbf{0.61 ± 0.06} & \textbf{0.94 ± 0.03} & \textbf{1.72 ± 0.04} \\
\bottomrule
\end{tabular}
}
\label{tab:navi_comparison}
\end{table}

\subsection{Navigation Trajectory Comparison}

\begin{figure}[ht]
  \centering
  \begin{subfigure}[t]{0.48\columnwidth}
    \centering
    \includegraphics[width=0.8\linewidth,height=4.0cm]{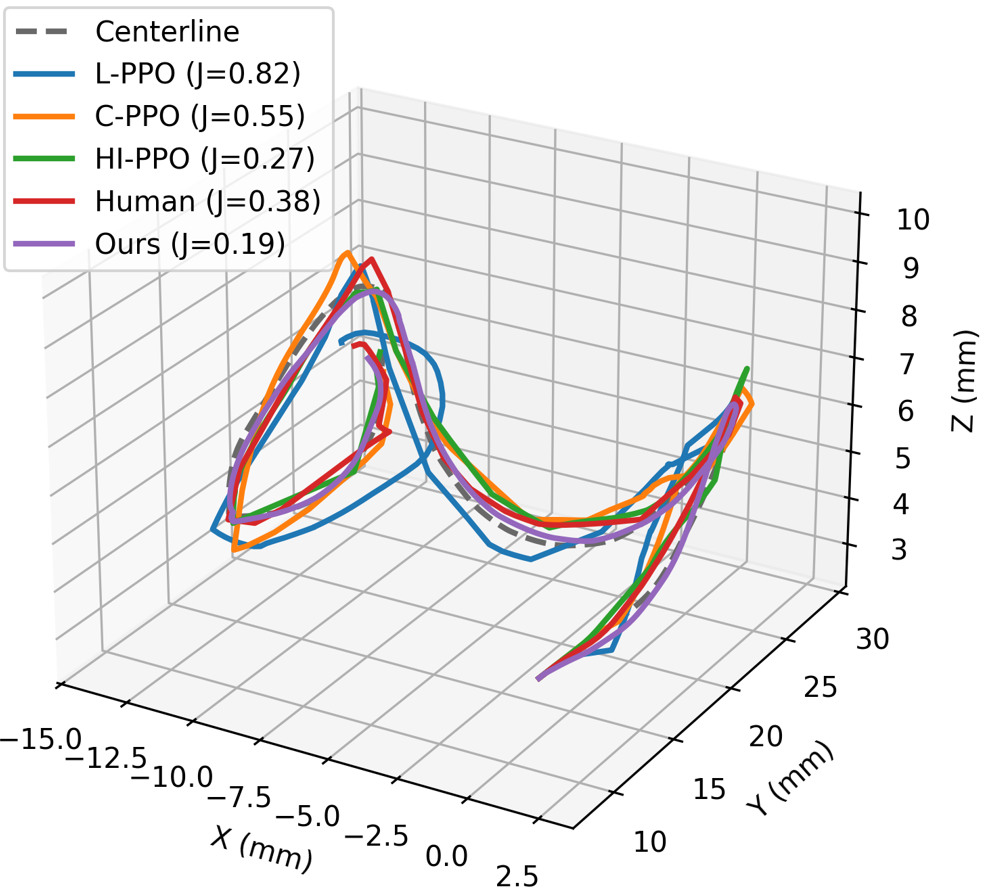}
    \caption{3D Navigation Trajectories.}
    \label{fig:nav_traj_3d}
  \end{subfigure}\hfill
  \begin{subfigure}[t]{0.48\columnwidth}
    \centering
    \includegraphics[width=0.8\linewidth,height=4.2cm,keepaspectratio]{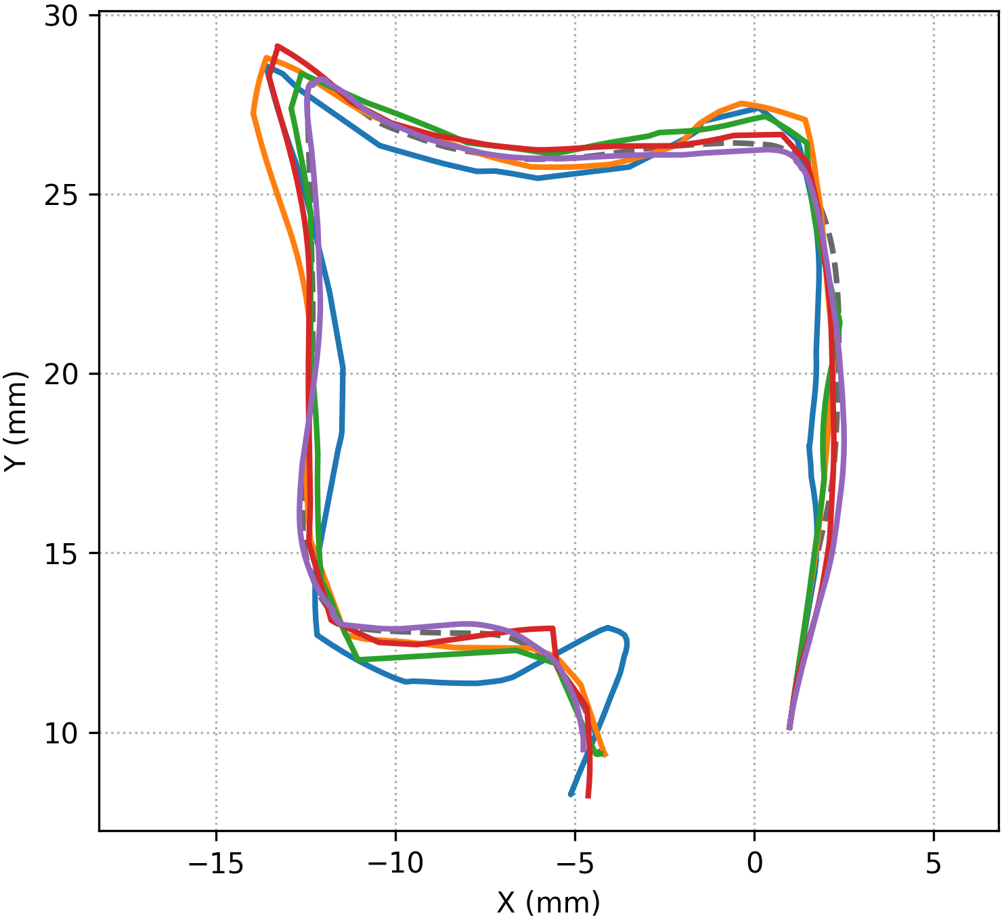}
    \caption{2D View of Trajectories.}
    \label{fig:nav_traj_topdown}
  \end{subfigure}
  \caption{Visualization of navigation trajectories in colon environment $C_2$ obtained by different methods. (a) shows the 3D trajectories, and (b) presents their top-down projections.}
  \label{fig:nav_traj_all}
\end{figure}

We conducted a comparative visualization of navigation trajectories in the colon environment $C_2$, 
where a human expert manually controlled the endoscopic robot for reference. 
As shown in Fig.~\ref{fig:nav_traj_all}, both our method and \textit{HI-PPO} outperform the human expert. 
Notably, our approach adheres more closely to the anatomical centerline, ensuring safe traversal and minimizing the risk of wall contact that could cause patient discomfort. 
In contrast, \textit{L-PPO} and \textit{C-PPO} exhibit pronounced oscillations, which can be attributed to their inaccurate depth estimation. 
By fine-tuning the foundation depth estimation model, our method effectively mitigates localization errors and achieves smoother and safer navigation.

\subsection{Ablations on DepthColNet}
To investigate the contribution of each component in the proposed \textit{DepthColNet}, we conducted an ablation study on the \textit{EndoSLAM} dataset. Compared with the baseline model, our method achieves a $3.9\%$ improvement in $\text{Abs. Rel.}$ and a $2.9\%$ gain in $\delta_1$. Specifically, we analyze the effects of the fine-tuning adapter and the proposed Colon Depth Enhancement Block (CDEB). As shown in Table~\ref{tab:ablation_depth}, both components contribute positively to improving depth estimation accuracy. The adapter achieves effective domain adaptation through low-rank feature transformation, while the CDEB refines depth prediction using depthwise convolution and channel attention. This structure captures intestinal contours and depth gradients with fewer parameters, enhancing sensitivity to subtle geometric variations while preserving computational efficiency.

\begin{table}[ht]
\centering
\tiny
\caption{Ablation of DepthColNet on EndoSLAM dataset.}
\renewcommand{\arraystretch}{1.2}
\resizebox{\columnwidth}{!}{%
\begin{tabular}{lcc}
\toprule
\textbf{Configuration} & \textbf{Abs. Rel. ($\downarrow$)} & \boldmath{$\delta_1$ ($\uparrow$)} \\
\midrule
Baseline (Depth Anything) & 0.284 & 0.648 \\
w/ Adapter only & 0.258 & 0.663 \\
w/ CDEB only & 0.267 & 0.659 \\
\textbf{w/ Adapter + CDEB (Ours)} & \textbf{0.245} & \textbf{0.677} \\
\bottomrule
\end{tabular}
}
\label{tab:ablation_depth}
\end{table}

\subsection{Ablations on DRL Reward Components}
To evaluate the contribution of each reward component in the proposed deep reinforcement learning framework, we conducted ablation studies in the colon environment \(C_1\), as summarized in Table~\ref{tab:ablation_reward}, because its sufficiently complex geometry better highlights the importance of different reward mechanisms. Starting with the distance-based reward $r_{\text{dis}}$, which aligns the navigation target with the image center, we progressively incorporated additional reward terms. The inclusion of the directional consistency term $r_{\text{dir}}$ significantly improved spatial alignment (reducing $S_{\text{geo}}$ by $0.64$), demonstrating more precise motion control along the lumen direction. The success reward $r_{\text{succ}}$ further enhanced navigation stability and improved $S_{\text{nav}}$ by $0.12$, while the stability term $r_{\text{Stability}}$ effectively suppressed oscillations. Finally, adding the step penalty $r_{\text{step}}$ achieved the best overall performance, reducing $J$ by $0.8$. These results confirm that each reward component contributes synergistically to stable, accurate, and geometry-aware navigation.

\begin{table}[ht]
\centering
\tiny
\caption{Ablation on reward components in colon $C_1$.}
\renewcommand{\arraystretch}{1.15}
\resizebox{\columnwidth}{!}{%
\begin{tabular}{lccc}
\toprule
\textbf{Rewards+} & $S_{\text{geo}}$ ($\downarrow$) & $S_{\text{nav}}$ ($\uparrow$) & $J$ ($\downarrow$) \\
\midrule
+ $r_{\text{dis}}$        & 1.47 ± 0.22 & 0.42 ± 0.11 & 3.15 ± 0.31 \\
+ $r_{\text{dir}}$        & 0.83 ± 0.03 & 0.74 ± 0.01 & 2.71 ± 0.11 \\
+ $r_{\text{succ}}$       & 0.76 ± 0.11 & 0.86 ± 0.08 & 2.52 ± 0.34 \\
+ $r_{\text{Stability}}$    & 0.62 ± 0.10 & 0.92 ± 0.14 & 2.11 ± 0.26 \\
+ $r_{\text{step}}$       & \textbf{0.61 ± 0.06} & \textbf{0.94 ± 0.03} & \textbf{1.72 ± 0.04} \\
\bottomrule
\end{tabular}}
\label{tab:ablation_reward}
\end{table}

\begin{figure}[ht]
  \centering
  \begin{subfigure}[t]{0.48\columnwidth}
    \centering
    \includegraphics[width=\linewidth,height=2cm]{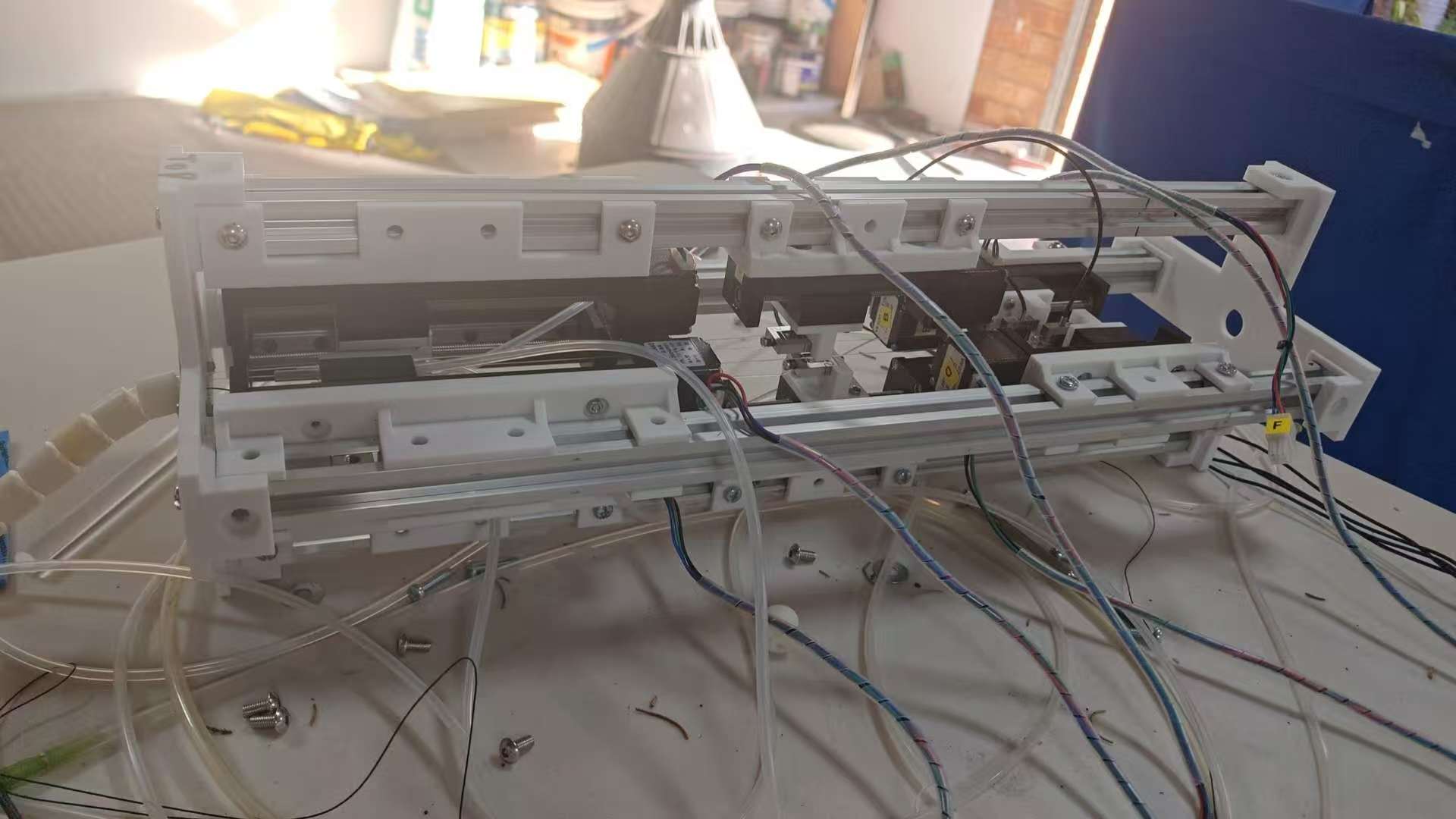}
  \end{subfigure}
  \hfill
  \begin{subfigure}[t]{0.48\columnwidth}
    \centering
    \includegraphics[width=\linewidth,height=2cm]{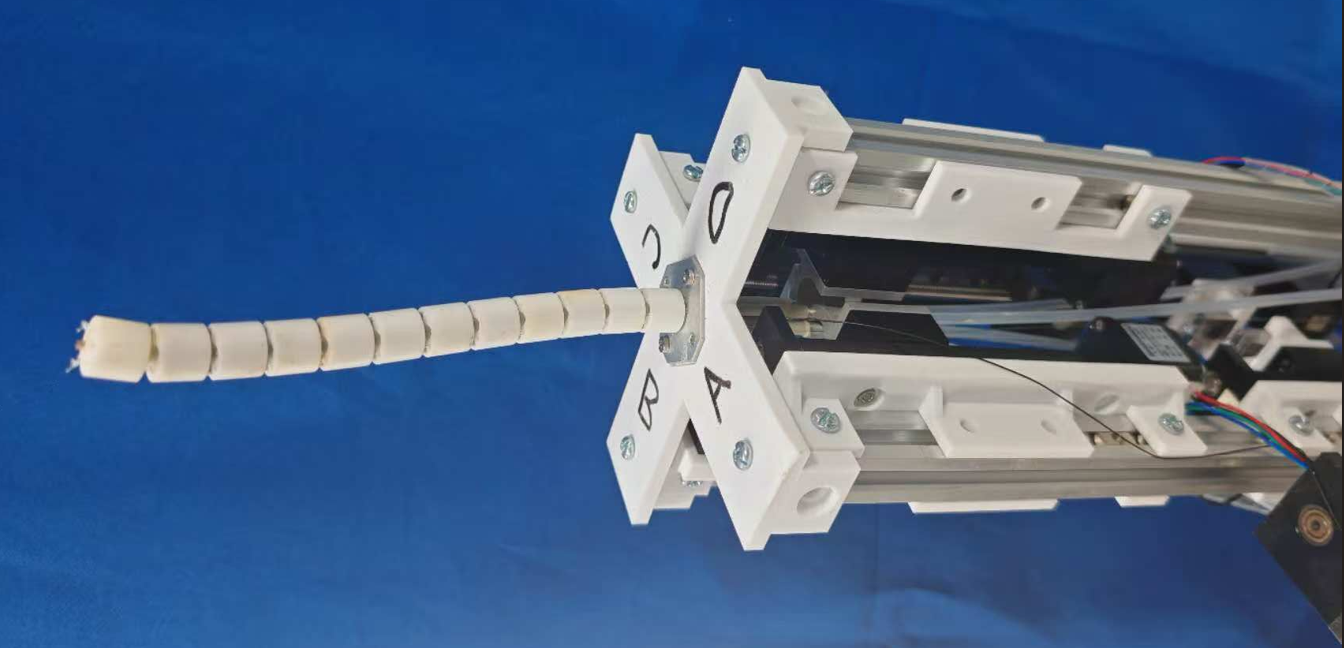}
  \end{subfigure}
  \caption{Physical prototype of the Follow-the-Leader (FTL) endoscopic robot used for sim-to-real validation.}
  \label{fig:real_rot}
\end{figure}

\vspace{-10pt}

\section{Conclusions and Future Work}
This paper presents an autonomous navigation framework for a follow-the-leader endoscopic robot that integrates vision-based perception with deep reinforcement learning. An intestinal simulation was built in \textit{NVIDIA Omniverse} for training and evaluation. By fine-tuning a foundation depth model on synthetic and real data, the robot achieved reliable monocular depth estimation and geometry-aware navigation. Experiments show that our method enables safe, stable, and centerline-consistent navigation in complex colon environments.

However, there are also limitations. First, the colon environment adopted in this experiment considers diverse textures and complex trajectories; however, the internal grease level has not been modeled. In real-world scenarios, the amount of intestinal grease varies among individuals of different body types, which may interfere with depth estimation due to specular reflections. 
In future work, we plan to design more diverse and realistic environments for image synthesis and testing. Second, the current experiments were conducted entirely in a simulation environment, 
where our approach achieved promising results. 
We are actively addressing the \textit{sim-to-real} gap and have already developed a physical prototype of the endoscopic robot, as shown in Fig.~\ref{fig:real_rot}. 
In future work, we plan to deploy our autonomous navigation algorithm onto the real robot 
and further tackle challenges such as signal transmission and latency.


\ifCLASSOPTIONcaptionsoff
  \newpage
\fi

\small
\bibliographystyle{IEEEtran}
\bibliography{citations}

\end{document}